\pdfoutput=1
\documentclass[11pt]{article}

\usepackage[margin=1in]{geometry}
\usepackage[utf8]{inputenc}
\usepackage[T1]{fontenc}
\usepackage{lmodern}
\usepackage{hyperref}
\usepackage{url}
\usepackage{xurl}
\usepackage{booktabs}
\usepackage{amsfonts}
\usepackage{nicefrac}
\usepackage{microtype}
\usepackage{xcolor}
\usepackage{comment}
\usepackage{wrapfig}
\usepackage{multirow}
\usepackage{subcaption}
\usepackage{graphicx}
\usepackage{float}
\usepackage{amsmath}
\usepackage{amssymb}
\usepackage{arydshln}
\usepackage{array}
\usepackage{tabularx}
\usepackage{natbib}
\usepackage{etoc}

\usepackage[capitalize,noabbrev]{cleveref}
\usepackage{mathtools}
\usepackage[inline]{enumitem}
\usepackage{multirow}
\usepackage{tabularx}

\crefname{equation}{Eq.}{Eqs.}
\crefname{figure}{Fig.}{Figs.}
\crefname{section}{Sec.}{Secs.}
\crefname{subsection}{Sec.}{Secs.}
\crefname{theorem}{Thm.}{Thms.}
\crefname{appendix}{Appx.}{Appx.}
\crefname{lemma}{Lemma}{Lemmas}
\crefname{algocf}{Alg.}{Algs.}
\Crefname{algocf}{Algorithm}{Algorithms}
\hypersetup{
  colorlinks,
  linkcolor={blue},
  citecolor=blue,
  urlcolor=blue,
  breaklinks=true,
  linktoc=all,
}

\usepackage{titlesec}
\titlespacing*{\section}{0pt}{*1.2}{*0.6}
\titlespacing*{\subsection}{0pt}{*1}{*0.4}
\titlespacing*{\subsubsection}{0pt}{*0.8}{*0.3}

\usepackage{amsmath,amsfonts,bm}

\def\eqref#1{equation~\ref{#1}}

\def\1{\bm{1}}

\DeclareMathAlphabet{\mathsfit}{\encodingdefault}{\sfdefault}{m}{sl}
\SetMathAlphabet{\mathsfit}{bold}{\encodingdefault}{\sfdefault}{bx}{n}

\renewenvironment{wrapfigure}[3][]{\begin{figure}[t]\centering\begin{minipage}{#3}}{\end{minipage}\end{figure}}

\newif\ifcomments
\commentsfalse
\ifcomments\newcommand{\comments}[1]{#1}\else\newcommand{\comments}[1]{}\fi
\definecolor{red}{rgb}{.8,0,0}
\definecolor{blue}{rgb}{0,0,0.8}

\title{Benchmarks as Software: Terminal-Bench and the Case for Continuous Benchmarks}
\author{%
  E. Kelly Buchanan$^{1}$\thanks{Correspondence: \texttt{kelly.buchanan@stanford.edu}} \quad
  Crystal Zhou$^{2}$ \quad
  Scott W. Linderman$^{1}$ \quad
  Christopher R\'e$^{1}$ \\[4pt]
  {\normalsize $^{1}$Stanford University \quad $^{2}$Cornell University}%
}
\date{}

\begin{document}

\maketitle
\etocdepthtag.toc{main}

\begin{abstract}
Coding agents are among the most economically consequential deployments of LLMs to date, making reliable evaluation increasingly important. Yet even benchmarks that are carefully validated at release can become stale. As models improve, they can discover valid solutions that existing verifiers reject, while changes to evaluation infrastructure can invalidate assumptions embedded in tasks and tests. We call the resulting accumulation of defects \textit{benchmark maintenance debt}: a growing gap between benchmark scores and the capabilities they are intended to measure.

We introduce a continuous-validation workflow for agentic benchmarks that combines community reports, task artifacts, agent traces, automated oracle checks, and human-gated repairs. We apply this workflow to Terminal-Bench~2.0, a benchmark that underwent extensive human and LM-assisted review and red-teaming before release. Our audit identifies issues in roughly 30\% of the benchmark, and we release the resulting repairs as Terminal-Bench~2.1. These repairs change evaluation results: accuracy increased by over 10 percentage points and leaderboard positions move by up to three ranks, even as the overall ranking remains highly correlated (Spearman $\rho = 0.958$).

Beyond accuracy, we evaluate how resource allocation shapes benchmark conclusions. We show that fixed wall-clock and output-token budgets can materially change apparent agent performance: For example, extending the time limit closes a 20~pp harness gap on GPT~5.4 and shrinks a 30~pp gap on GPT~5.4 mini to 5~pp. Benchmark maintenance is therefore not only about repairing task artifacts: resource budgets are part of the measurement specification and must be validated alongside the tasks themselves.

More broadly, our results highlight that benchmark maintenance debt is not just accumulated bugs: benchmarks sit in an evolving ecosystem of dependencies, infrastructure, and agents, so their assumptions and resource settings must be revalidated over time.

\end{abstract}

\section{Introduction}
\label{sec:introduction}

\begin{figure}[t]
  \centering
  \includegraphics[width=\linewidth]{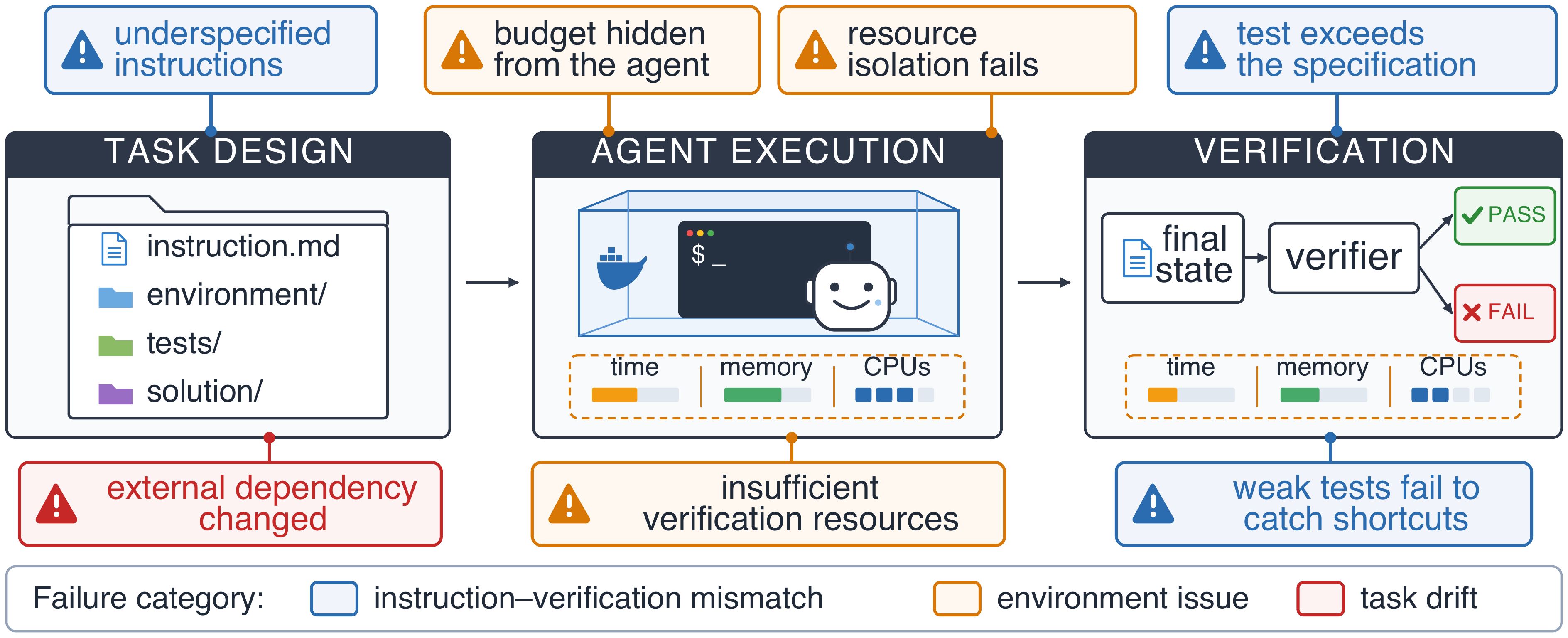}
  \caption[Failure surfaces in agentic-benchmark tasks.]{%
    \textbf{Failure surfaces along the agentic-task lifecycle}. From authoring to execution and verification, each stage can admit a failure that contributes to benchmark maintenance debt. Colors mark the three failure categories (blue: 
    instruction--verification mismatch, orange: environment issues, red: task drift).
  }
  \label{fig:teaser}
\end{figure}

Agentic benchmark scores shape product claims and research investment. But agentic benchmarks are not static datasets: %
over time, these systems accumulate defects that decouple a model's score from the capability the benchmark is meant to measure. In turn \textit{benchmark maintenance debt}, the backlog of task-side defects that must be repaired or retired for benchmarks to remain valid. These surface along the agentic-task lifecycle as shown in \cref{fig:teaser}: 
 benchmark designers can encode assumptions about how agents will solve a task, which break when agents find valid solution paths the designer did not anticipate, or shortcuts that pass the tests without solving the task. Furthermore, tasks that depend on external state (packages, APIs, websites) can break when that state changes after release, causing failures unrelated to agent capability.

For agentic benchmarks, validity requires ongoing maintenance. Terminal-Bench~2.0~\citep{terminalbench} illustrates this need: its tasks underwent human review, LM-assisted auditing, and exploit testing before release~\citep{zhu2025abc}, yet within months its issue trackers contained over 50 open reports tagged as task defects~\citep{tb-repo,tb2-repo,zeng2026glm5}, spanning different assumptions made across the full task stack (Fig.~\ref{fig:teaser}). Extensive release-time validation therefore does not eliminate the need for subsequent auditing and repair.
Furthermore, a defect report does not establish its cause or justify a particular fix. Proposed repairs require their own validation. For example, we found some proposals addressed the underlying validity failure by clarifying the task or repairing the oracle; others only patched the observed symptom, for example by relaxing a regex or broadening an acceptance condition. Maintainers therefore need a consistent process for establishing defects, reviewing proposed changes, and checking that repairs preserve the intended task.

Resource allocation and infrastructure variation further complicate this process. Differences in time and memory limits, environment configuration, and model or sandbox providers can make the same task behave differently across evaluation setups, complicating both the reproduction of reported failures and the validation of repairs~\citep{anthropic-infra-noise,openai2025ghosts}.
Freezing environments trades fidelity for reproducibility, filtering questionable tasks~\citep{swebench-verified} trades coverage for precision, and refreshing evaluation sets~\citep{jain2024livecodebench}  keeps tasks unseen by models but does not fix tasks that break after release.
We develop a complementary benchmark audit tool for reviewing tasks. We apply it to Terminal-Bench~2.0, release the resulting update as Terminal-Bench~2.1, and examine how the revisions affect evaluation results.\par

\begin{figure*}[t]
  \centering
  \captionsetup{skip=3pt}
  \includegraphics[width=\linewidth]{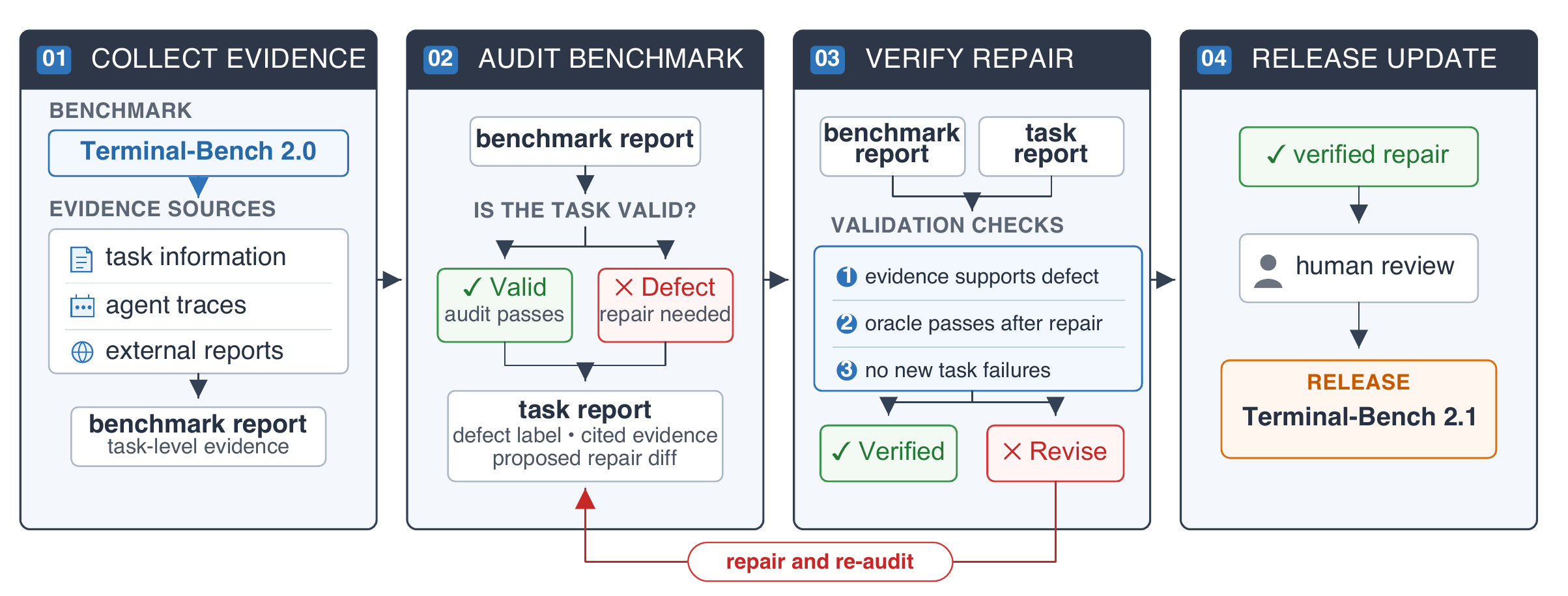}
  \caption[TB-Audit Continuous-validation workflow.]{%
    \textbf{Agentic pipeline for continuous benchmark validation.}
    An evidence collector agent gathers complementary evidence streams: static task files,
    prior agent traces, and external reports. An auditor agent applies audit principles
    for specification fidelity, tests, resources, and risk, then drafts a cited repair, which is checked against
    evidence while \texttt{/oracle} rebuilds the task and runs the reference solution. The
    repair workflow can open a draft PR with evidence; maintainer review gates the verified release.}
  \label{fig:continuous_validation}
\end{figure*}

\noindent
Our contributions are:
\begin{itemize}[itemsep=0pt]
\item \textbf{A continuous validation workflow.} We show that benchmark validation can itself be continuous: modern agentic tools make it practical to combine community reports, public execution traces, agent-assisted triage, and human adjudication into a maintainable workflow for identifying, repairing, retiring, and documenting benchmark issues %
(\Cref{sec:pipeline}).
\item \textbf{An updated benchmark, Terminal Bench 2.1.} We apply this workflow to Terminal-Bench~2.0 and release Terminal-Bench~2.1. The workflow surfaces issues in roughly one third of the benchmark by combining community reports, agent-assisted review, and maintainer adjudication. We categorize these issues into three recurring sources of benchmark maintenance debt: \emph{environment issues}, \emph{instruction--verification mismatches},  and \emph{task drift} (\cref{sec:failure_categorization}).
\item \textbf{The effects of benchmark repair.} We re-evaluate 19 agents on Terminal-Bench~2.0 and Terminal-Bench~2.1 and show that repair can change reported accuracy and efficiency conclusions without fully overturning the leaderboard. Agent rankings remain highly correlated, but individual scores move by nontrivial margins (\cref{sec:results}). Even with feasibility restored, we highlight the changes in performance due to resource allocation (e.g.~>50\% of failed trials are due to timeouts, and up to $\sim$68\% for some harnesses) (\cref{fig:figure2_failures}b).
\end{itemize}

\section{Related Work}
\label{sec:related_work}

\paragraph{Benchmark auditing.}
Prior work has studied how benchmark conclusions change when the evaluation
set is revised. \citet{recht2019imagenet} resampled ImageNet and CIFAR test sets and used rank correlation between the original and new test sets to evaluate whether a benchmark measures a stable construct. We adopt the same protocol, and apply it to a different failure mode: task-side defects in an executable agent benchmark.

A second response is to validate and filter benchmark instances.
 SWE-Bench Verified~\citep{swebench-verified} contracted professional developers to audit SWE-Bench~\citep{swebench} and released a conservatively filtered subset. WebArena Verified~\citep{webarena-verified} applies a similar validated-subset approach to WebArena~\citep{webarena}. Related audits identify solution leakage, weak test suites, and task/outcome-validity failures in agentic benchmarks
\citep{aleithan2024swebenchplus,yu2025utboost,zhu2025abc}. These approaches typically treat validation as a one-shot curation pass and reduce the benchmark rather than repairing it.

A third response is to time-window or refresh benchmarks.
LiveCodeBench~\citep{jain2024livecodebench} and LiveBench~\citep{livebench}
continuously add new problems to reduce contamination and benchmark staleness. This approach is closest in spirit to continuous validation, but instead we focus on repairing rather than filtering.

\paragraph{Auditing tooling.}
Benchmark auditing is facilitated by a growing stack of operational tools. The Agentic Benchmark Checklist (ABC)~\citep{zhu2025abc} provides rubrics for auditing outcome validity and task validity. Tools such as Docent~\citep{meng2025docent} and LangSmith \cite{langsmith} make large-scale inspection tractable by ingesting benchmark trajectories for review. Complementary adversarial tools audit benchmark hackability by searching for ways that agents can pass tests without satisfying the intended task. These tools address different parts of the audit pipeline. Rubrics specify the evaluation criteria; ingestion tools collect evidence; red-teamers identify candidate defects; and human adjudication resolves ambiguous cases. We combine these ingredients into a continuous validation workflow (\Cref{sec:methods}).

\paragraph{Agentic evaluation.}
Executable agent benchmarks do not evaluate a model in isolation: scores also depend on the agent harness, the execution environment, and the resource budget. Harness effects are well documented \citep{yang2024sweagent,wang2025openhands}. \citet{terminalbench} found that, on average, the choice of
harness matters less than the choice of model.
Environment effects have received less attention:
\citet{anthropic-infra-noise} shows that infrastructure configuration
alone can shift agentic coding scores. We extend this line of work
in two ways: averages can hide large within-vendor harness
gaps (${\sim}20$~pp for some models), and that these gaps largely reflect speed under the default time budget.

\section{Problem Setup}
\label{sec:problem_setup}
\label{sec:preliminaries}

\textbf{Every task has a failure surface.} A task is a timed exam with the following components:

\begin{itemize}[noitemsep]
  \item \textbf{Instruction:} the natural-language problem description shown to the agent.
  \item \textbf{Environment:} a Docker image with pre-installed software and any provided files.
  \item \textbf{Oracle:} the known-correct solution. %
  \item \textbf{Test:} the grading code that runs after the agent stops.
  \item \textbf{Resources:} CPU, memory, and wall-clock budgets.
\end{itemize}

An \emph{agent} is a model with access to tools.  The \emph{agent harness} is the system that gives a model the tools it needs to interact with its environment.  A \emph{trial} is one independent attempt at a task by an agent. For agent $a$ working on task $t$ with $n_{a,t}$ trials, $\hat p_{a,t}$ is the mean reward. The coverage, or \emph{pass@k}, estimates the probability that at least one of the trials succeeds. The coverage is the unbiased pass@k estimator from \citealt{chen2021codex} throughout. 

\paragraph{Datasets.}
We operate with two datasets: the \emph{evidence data} and the
\emph{evaluation data}. The evidence data is the external
evidence gathered on the benchmark, including trial data from the leaderboard, used to identify and fix
defective tasks and release Terminal Bench 2.1. The evaluation data is a set of TB2.0 and TB2.1 trials, run for 19 agents, used to measure the effect of the fixes.

To isolate the role of efficiency, we run an additional condition on TB2.0 and TB2.1 that deliberately deviates from the benchmark
guidelines: we increase the resources allocated, giving agents more
time to complete each task. Owing to cost, this condition is run on a subset of the agents.

\section{Benchmark Auditing Workflow}
\label{sec:methods}
\label{sec:pipeline}

Our goal is to audit agentic benchmarks for task-side defects and to produce candidate repairs that maintainers can review. We have at our disposal three evidence streams: the task's specification, prior execution traces, and community reports from the web. We want this evidence compiled into a task-level audit report: a defect label, supporting evidence, and a proposed fix when one is available. We define three agent workflows for collecting evidence, auditing, and verifying the final report. 

\paragraph{Audit rubric.}
A central requirement of the workflow is a rubric. Without one, an agent can flag suspicious behavior, but its behavior may not be consistent or easy to parse.
Previous work has developed rubrics to audit benchmarks \citep{zhu2025abc}. We build on this idea, but adapt it for a maintenance workflow. In this setting, we found it more useful to ask for a small number of action-oriented defect labels tied to likely repairs, rather than broad free-form validity judgments. The auditor therefore assigns a primary task-side defect category and records secondary labels when applicable. This makes reports comparable across agents and gives maintainers a concrete review target. 
We ask auditors to classify each task using the following primary \emph{defect categories}:
\begin{itemize}
\item \textbf{No defect found:} the task is well specified, the oracle is accepted, the verifier checks the intended behavior, and the stated resource budget is sufficient.
\item \textbf{Instruction--verification mismatch:} the task instruction and the verifier disagree. This includes underspecified instructions whose hidden constraints appear only in tests, as well as overspecified tests that reject behavior satisfying the task as written.
\item \textbf{Task drift:} the task changed after release because of mutable external state, such as package versions, network resources, public APIs, or web content.
\item \textbf{Environment issues:} the task is solvable in principle, but the provided environment and resources (CPU, memory, storage, or system/network configuration) prevent the oracle, the verifier, or a valid agent solution from running reliably.
\end{itemize}

\paragraph{Audit workflow.}
The \texttt{audit} workflow is illustrated in \Cref{fig:continuous_validation}. For each task, the \emph{evidence} agent produces a structured report containing information on the benchmark, the task specification, trace information, and community reports. The role of this agent is to organize available evidence for a downstream auditor, while keeping observation separate from adjudication.

An \emph{auditor} applies the rubric to each task. The auditor checks whether the task is well specified, solvable, verifiable, and whether the verifier accepts valid solutions and rejects invalid ones. The rubric here follows the ABC guidelines \citep{zhu2025abc}, and it produces a task-level report containing a primary defect label, a short justification, the location of the defect, and a proposed repair.

A \emph{verifier} reviews the evidence and the auditor's label. We include this stage because auditing alone did not capture task drift reliably in the pilot runs. The verifier agent is asked to run the oracle, challenge the proposed label, identify missing evidence, and check whether the proposed fix addresses the underlying defect. The full breakdown is in \Cref{app:audit_pipeline}.

\begin{figure}[t]
  \centering
  \captionsetup{skip=3pt}
  \includegraphics[width=\linewidth]{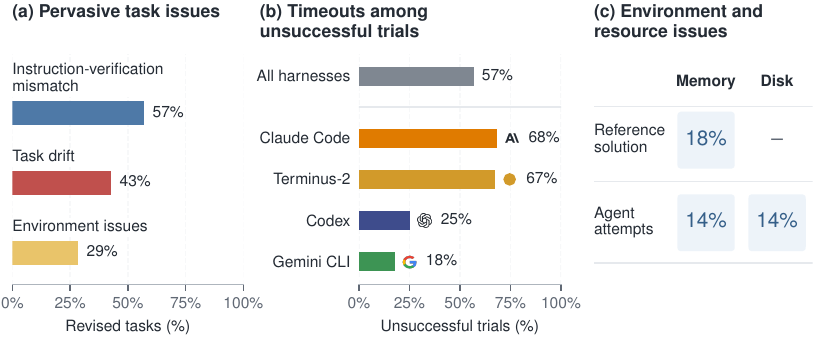}
  \caption{%
    \textbf{Three views of where task errors come from.}
    (a) Prevalence of defect categories in 28 tasks from Terminal Bench 2.0; a task can have more than one.
    (b) Prevalence of timeouts among failed Terminal Bench 2.1 trials at the default time budget, by harness (the set of agents varies by harness).
    (c) Share of revised tasks where the reference solution or agent attempts hit or reached the Terminal Bench 2.0 memory or disk limit.
  }
  \label{fig:figure2_failures}
\end{figure}

\paragraph{Repair conventions.}
Once a defect is identified, the repair is not always mechanical, but involves human judgment~\citep{zhu2025abc}. We follow these conventions for each defect category:
\begin{itemize}
\item \textbf{Instruction--verification mismatch}: any solution satisfying
the task as written is accepted, and invalid solutions are rejected.
\item \textbf{Task drift}: make the dependency reproducible where possible and remove any dependency that prevents the oracle from running reliably.   
\item \textbf{Environment Issues:} We set a \emph{feasibility floor}: a known-correct oracle solution and the verifier should run reliably within the benchmark budget. 
\end{itemize}

\paragraph{Resource issues.}
For resource-related failures, we adjust the binding resource axes (wall-clock time, memory, or CPU count; sometimes more than one) to the factor closest to the original proposal or at least $1.5\times$ the observed requirement of the oracle solution and verifier. 

This convention serves two purposes. First, it prevents false negatives caused by infrastructure noise: a correct solution should not fail because the verifier or oracle only barely fits within the budget, but we should leave enough slack for the problem to be solved. Second, the resource sets the task difficulty, and here we leave those as close to what the original task contributors intended.  

An alternative would be to tune resource budgets per task. However, doing so would require repeated sweeps over any resource and would introduce an additional subjective design choice into the process. While the feasibility rule is conservative, it is an auditable convention. %

\paragraph{Shared repair rules.} When tasks depend on live APIs, the difficulty can change after the audit. When a repair changes task difficulty, we document the change rather than tuning the task back to match its previous pass rate by following expert guidelines.  

\paragraph{Workflow evaluation.}
We evaluate the audit workflow in the setting it was designed for: post-release benchmark maintenance. We used the workflow during Terminal-Bench~2.0 maintenance to surface candidate task defects, attach evidence, and propose repairs for maintainer review. The accepted repairs were then shipped as Terminal-Bench~2.1. 

We evaluate the workflow along three dimensions. First, we ask whether the workflow is agent-agnostic: under the same rubric, do different auditors assign similar defect labels? Second, we ask whether this agreement persists when the workflow is instantiated with lower-cost auditor models. Third, we evaluate the workflow as a maintainer-in-the-loop triage system by comparing its flags against a human-adjudicated set of repairs.

As we described previously, the agents assign a task to a defect category. The defect categories were selected based on results from a 15-task pilot subset of Terminal-Bench~2.0 and approximately 1,500 TB2.0 execution trials. The pilot fixed only the category definitions, not per-task labels.

To evaluate whether the workflow is agent-agnostic, we ran it with two classes of auditors.  Managed coding assistants, Claude Code (Opus~4.6) and Codex (GPT-5.4),
agree on the primary defect category, including \emph{no defect}, at
$\kappa=0.80$ over all 89 TB2.0 tasks.

Lower-cost API auditors agree on the binary triage decision (flagged vs. clean) at $\kappa = 0.55$--$0.77$ over the same 89 tasks (\Cref{tab:canary_agreement}), but less on attribution: per-category agreement (one-vs-rest, since a task can carry multiple labels) is moderate for instruction--verification mismatch ($\kappa=0.60$) and environment issues ($\kappa=0.59$), and low for task drift  ($\kappa=0.21$--$0.41$). We therefore rely on the deterministic oracle check rather than auditor labels to detect drift.  Rerunning the reference solution helps detect drift that disrupts the reference execution or its verification.

These results suggest that capable auditors are largely agent-agnostic in both triage and attribution, while lower-cost auditors are reliable for triage but require the oracle check and maintainer review for attribution. Agreement measures consistency rather than correctness. An extended breakdown appears in \Cref{app:audit_pipeline}.

To prevent leakage from the community reports the workflow is designed to consume, all evaluation runs disable the auditors' web-search tools; in deployment, web search stays on.

Against the final TB2.1 release, the workflow reaches 57.1\% precision and 71.4\% recall. Because the workflow contributed to constructing the repair set, these metrics measure retrospective agreement with the release rather than independent defect discovery; the workflow is best viewed as a triage tool, the outputs of which are then filtered by maintainer review.

\section{From Terminal Bench 2.0 to 2.1}
\label{sec:failure_categorization}

Applying the audit workflow to Terminal-Bench 2.0 surfaced issues in 28 tasks, roughly a third of the benchmark. These consolidate defects found in prior audits, namely the 11 tasks flagged in Terminal-Bench Verified \citep{zai-tb2-verified} (compared in \cref{app:tb_zai}) and 5 further tasks addressed across various public issues and pull requests \citep{tb-repo,tb2-repo}, together with 14 additional tasks that the audit newly identified as needing repair. Two tasks can appear in more than one source, giving 28 in total.

\cref{fig:figure2_failures} shows the breakdown of the task defects in the 28 tasks that needed repair. The most common failure mode was instruction--verification mismatch (present in 57\% of the tasks). For example, on the task \texttt{query-optimize}, the verifier enforced byte-for-byte preservation of the database file, a requirement absent from the original instructions.  The repair made this restriction explicit.  \citep{aleithan2024swebenchplus,yu2025utboost}.

The second most common defect was task drift (43\% of tasks). Tasks that depend on external state, such as package versions, public APIs,
and web content, break when that state changes after release. We repair
drift by removing the dependency where possible and pinning it
otherwise (See \cref{app:audit_pipeline}).

The third category, environment issues (29\% of tasks), comes from
resource budgets. In TB2.0, task contributors set CPU, memory, and time
budgets for each task based on their own resource use. These
estimates were sometimes too tight (Fig.~\ref{fig:figure2_failures}c):
the oracle solution, the verifier, or the agent hit or reached its memory or disk limit, so trials could fail because of resource limits rather than
incorrect solutions. Execution backends (Docker, Daytona) also differ
in how they isolate filesystems, enforce memory limits, and expose
networks between agent and verifier. The audit raises the binding budgets so
the oracle runs with at least $1.5\times$ slack on each,
restoring a feasibility floor.

An audit of this size is dangerous: if repairs reorder the leaderboard, prior claims about relative agent performance may need to be revisited. We test this directly in the next sections.

\section{Results}
\label{sec:results}

 The performance of an agent is a product of the model, harness and 
 resources \citep{yang2024sweagent, kapoor2025hal, terminalbench}. While we generally compare model and harness, here we show the impact of resources as well. We use TB2.1 to ask which conclusions are stable under changes to this measurement stack. First, we compare TB2.0 and TB2.1 at the default protocol to test whether repairs change accuracy and rankings. Second, we vary the time budget to test whether repairs change the efficiency frontier: which agents solve tasks, and how much budget they need to do so. Third, we ask how much of a fixed-budget harness gap is explained by speed under the timeout. Scalar leaderboards hide these
distinctions because they collapse benchmark version, harness, and resource
budget into one number.

\subsection{The audit raises accuracy and changes close rankings}
\label{sec:ranking_stable}
\begin{textwrapfigure}{r}{0.55\linewidth}
  \vspace{-1.0\baselineskip}
  \centering
  \captionsetup{skip=3pt}
  \includegraphics[width=\linewidth]{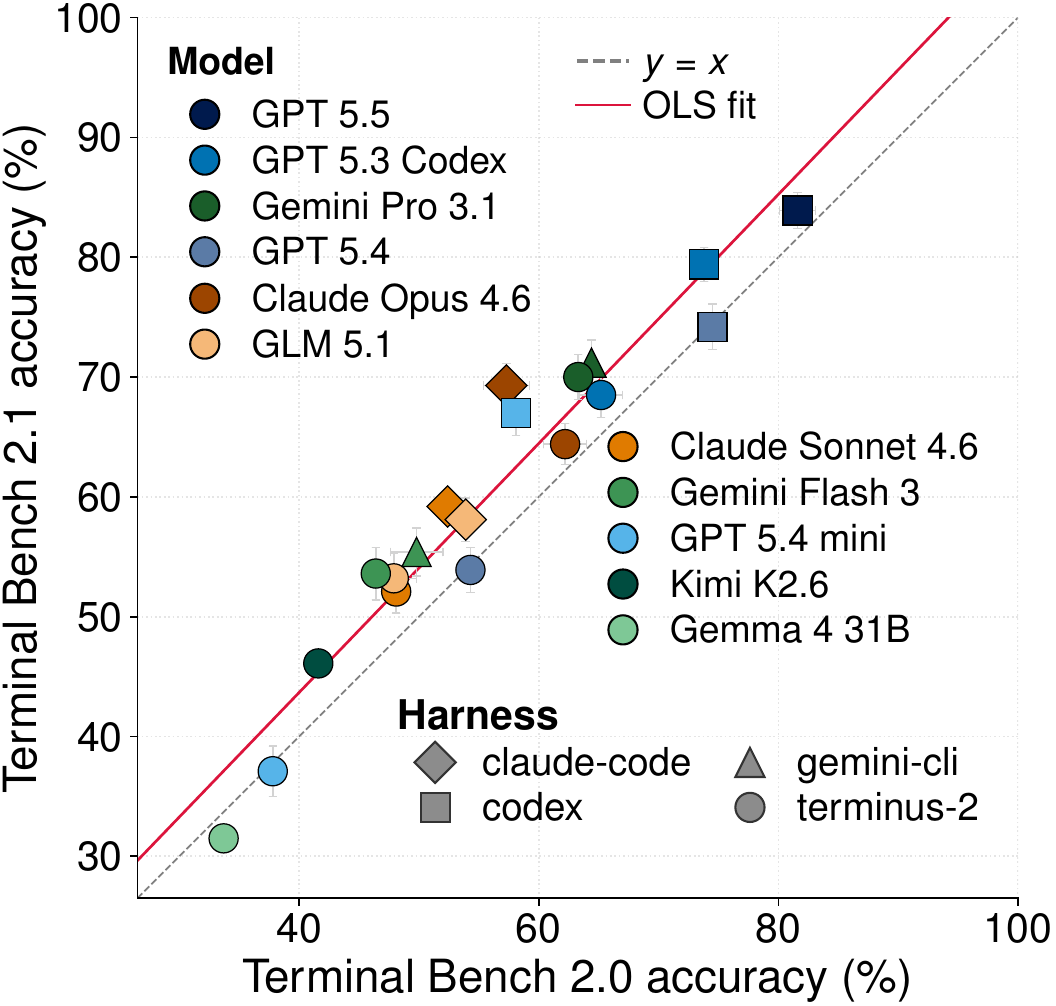}
  \caption{%
    \textbf{Terminal Bench 2.1 changes accuracy unevenly across agents.}
    Spearman $\rho = 0.958$ with a task-bootstrap 95\% CI of
    $[0.881, 0.979]$ across 19 agents.
  }
  \label{fig:leaderboard_scatter}
  \vspace{-0.6\baselineskip}
\end{textwrapfigure}

We evaluate 19 agents from multiple providers and compare the accuracy in both TB~2.0 and TB~2.1, as illustrated in \Cref{fig:leaderboard_scatter}. The correlation is high (Spearman $\rho = 0.958$), with an OLS fit of $\text{TB2.1}=2.1 + 1.04\,\cdot\,\text{TB2.0}$. The lift is uneven: Claude Code~+~Claude~Opus~4.6 lifts over $7$~pp above the fit and climbs three places, while Codex~+~GPT~5.4 sits below the fit. The two weakest agents, which fall below 40\% on TB~2.0 accuracy, both fall below the fit and lose accuracy on TB~2.1. This is consistent with the difficulty of the repaired tasks: 19 of the 28 repaired tasks were hard, 7 medium, and 2 easy, so repairs that make hard tasks solvable matter little to agents that rarely solve them.

The largest gains are concentrated among tasks with
instruction--verification mismatches. In TB~2.1, five tasks increased from pass rates of at most $15\%$ to over  $50\%$ after repairs centered on the verifier, accounting for roughly $62.7\%$ of the total absolute pass-rate change on the modified-task subset (full breakdown in \Cref{app:supp_results}). These tasks fall under instruction--verification mismatch: agents were often producing valid solutions, but TB~2.0 failed to recognize them. The remaining 23 modified tasks contribute smaller changes in both directions. In some cases, the audit made tasks stricter by removing false-positive credit or tightening the intended specification, lowering TB~2.1 pass rates relative to TB~2.0.

\subsection{The audit can shift the efficiency frontier}
\label{sec:compute_scaling}
\label{sec:efficiency_tradeoff}

Setting resources for executable agent benchmarks is non-trivial. A task score depends not only on whether an agent can solve the task, but also on whether it can solve it within the allocated resources. \Cref{fig:figure2_failures}b shows that over 50\% of failed trials ended due to timeout, ranging from 68\% under \texttt{claude-code} to 18\% under \texttt{gemini-cli}; for the weakest agent--model pairs, timeouts dominate (\cref{app:failure_composition}).

\begin{figure*}[t]
  \centering
  \includegraphics[width=0.95\linewidth]{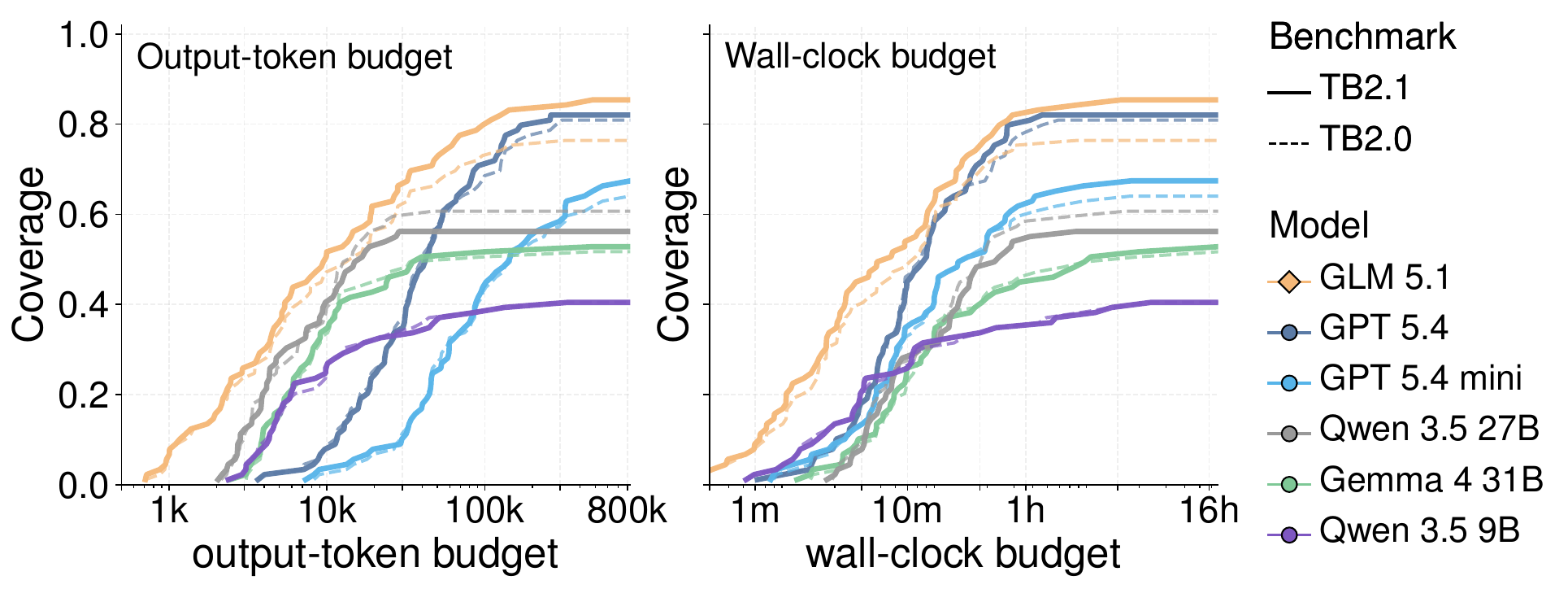}
  \caption{%
    \textbf{Cost-aware effectiveness shifts across the audit.} Coverage improves as resource budgets increase. %
  }
  \label{fig:eutb_eutimeb_scaling4}
\end{figure*}

\begin{textwrapfigure}{r}{0.45\linewidth}
  \vspace{-1.0\baselineskip}
  \centering
  \captionsetup{skip=3pt}
  \includegraphics[width=\linewidth]{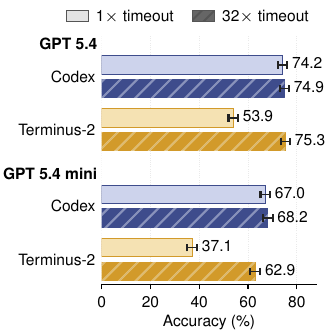}
  \caption{%
    \textbf{The harness gap is mostly a time-budget effect.}
    TB2.1 accuracy for Codex and Terminus-2 at the standard allocated time
    and with timeouts extended to $32\times$. Extending the budget closes
    most of the Codex--Terminus-2 gap.
  }
  \label{fig:harness_combined}
  \vspace{-0.6\baselineskip}
\end{textwrapfigure}

We therefore evaluate the test-time scaling properties of these models. For each agent--model pair, we increase the wall-clock budget up to $32\times$ the original task timeout and measure how many tasks are solved as the budget increases. This larger budget is an upper bound: agents can exit once they finish, so they do not need to consume the full allocation. Few timeouts remain at the largest budget, except for the small models included (Qwen~3.5), so most remaining failures are not timeout artifacts: these tasks stayed unsolved under the tested configurations.

\cref{fig:eutb_eutimeb_scaling4} shows coverage for six agent--model pairs: five use Terminus-2, while GLM~5.1 with Claude Code is included as a reference point because it lies on the efficient frontier. The solid curves show TB2.1 and the dashed curves show TB2.0. Each curve should be read in two ways. Its left--right position shows efficiency: a curve that rises earlier corresponds to an agent that solves tasks with less time or fewer output tokens. Its final height shows coverage at the largest budget: if the curve flattens below $1.0$, then giving more budget does not recover the remaining tasks.

First, agents differ substantially in efficiency. GLM~5.1 reaches high coverage at much smaller budgets than the other agents, so its curve sits farther to the left. GPT~5.4 reaches high final coverage as well, but at the expense of more tokens. The smaller open models, including Qwen~3.5~9B, Qwen~3.5~27B, and Gemma~4~31B, solve many tasks as the budget increases, but their curves flatten at lower coverage. Since the wall-clock sweep nearly eliminates timeout errors for most pairs, the remaining failures are not timeout artifacts: these tasks stay unsolved under the tested configurations.

Second, the audit changes these curves unevenly. GLM~5.1 solves 8 more tasks on TB2.1 and overtakes GPT~5.4, whose coverage barely changes. For the Qwen pairs, by contrast, the TB2.0 and TB2.1 curves are nearly unchanged. Thus, efficiency rankings can also be affected by the benchmark version.

\paragraph{Harness gaps depend on the time budget}
Prior work has shown that harnesses and agent-computer interfaces affect agent performance~\citep{yang2024sweagent, kapoor2025hal, terminalbench}. We ask: how much of a fixed-budget harness gap disappears when the slower harness is moved along the efficiency frontier?

\Cref{fig:harness_combined} uses within-vendor harness swaps as a diagnostic. At the default time, Codex gives much higher scores than Terminus-2 for OpenAI models. Harness choice moves GPT 5.4's accuracy by ${\sim}20$ pp, rivaling between-model performance. However, most of the harness gap is recovered when we give the tasks a longer budget, suggesting speed rather than capability gains.

The results here show that a fixed resource budget is an arbitrary operating point and that the choice is not innocuous: it can change efficiency conclusions for individual models. This supports setting a principled floor and testing sensitivity to the budget, rather than calibrating to any one choice of agents and models.

\section{External Validity: Auditing Harbor-Index v0.1}

To evaluate whether the workflow transfers beyond Terminal-Bench, we applied it to 34 randomly selected tasks from an 80-task snapshot during the development of Harbor-Index v0.1. The snapshot drew from 29 source benchmarks spanning diverse domains and evaluation procedures
\citep{shi2026harboradaptersharborindexinfrastructure}.
Many tasks had been reimplemented in a shared format, creating
additional opportunities for defects in instructions, environments,
and grading logic.

The Harbor team manually reviewed all 34 tasks, including those
the workflow did not flag. The workflow flagged 14 tasks, of which manual review confirmed 12 as defective (86\% precision).
Findings included instruction--verification mismatches,
infrastructure failures, and dependencies on mutable external
state. For example, several GAIA2 tasks rejected correct reference solutions because an auxiliary grading service was invoked under the wrong condition. In GSO, instructions omitted a condition required by the grader.

Manual review also identified defects the workflow missed:
an incorrect reference answer in Lab-Bench, nondeterministic
grading in QCircuitBench, and an HLE instruction admitting two
plausible correct answers. Two additional bugs in shared grading code affected 13 tasks, overlapping with tasks already flagged for other defects: the GSO grader executed the agent-produced program twice, and the GAIA2 grader mishandled special characters.

The results show that the workflow can identify defects beyond
Terminal-Bench. The missed cases motivate further investigation
of benchmark-specific checks, repeated execution, and domain
expertise. The Harbor-Index team subsequently extended validation with trajectory-level judges and further rounds of review and repair; we refer readers to
\citet{shi2026harboradaptersharborindexinfrastructure}
for details.

\section{Discussion}
\label{sec:conclusion}
\label{sec:discussion}

\paragraph{Continuous validation.}
The audit pipeline surfaced 28 task-side defects in TB2.0, including 12 cases of post-release drift. Drift is by definition a recurring failure mode: more dependencies will rot between this paper's submission and its publication. 

This is the case for version control rather than one-shot validation: each release should be a numbered dataset, pinned to task-level digests, with a public record of what changed. 
LiveCodeBench~\citep{jain2024livecodebench} already operates this way
for contamination control; the same discipline applies to task-side
defects. 

\paragraph{Ranking stability under benchmark repair}
Across 19 agents, rankings remain highly correlated between TB2.0 and TB2.1 (Spearman $\rho = 0.958$), while individual scores shift by up to 12 percentage points and leaderboard positions move by up to three ranks. Thus, the broad ordering is stable under these repairs, but absolute performance estimates and comparisons between nearby agents can depend on the benchmark version.

The largest gains concentrate on
instruction--verification mismatches, the most prevalent defect category. The audit also shows the sensitivity of efficiency conclusions to different design choices, and how setting a feasibility floor is an affordable fix.

Overall, this work provides (i) an audit methodology that surfaces and repairs benchmark defects; (ii) the TB2.1 artifact, demonstrating that audit-driven repairs can preserve agent rank ordering while shifting absolute scores; and (iii) evidence that, on TB2.1, both task definitions and resource budgets shape published scores in ways current reporting practice does not control for. %
Finally, because this pipeline can run continuously, Terminal Bench 2.1 is a checkpoint rather than a terminal state.

\subsection*{Reproducibility statement}

Terminal-Bench~2.1 is publicly available at 
\url{github.com/harbor-framework/terminal-bench-2-1}.  The task revisions and supporting discussion are documented in:
\url{github.com/harbor-framework/terminal-bench-2/pull/53}.

The audit workflow is described in \Cref{sec:pipeline}, and
\Cref{app:audit_pipeline} details the continuous-validation checks.
The evaluation setup is documented in \Cref{app:eval_setup},
including the models (\Cref{app:models}), harnesses
(\Cref{app:harnesses}), and per-agent configurations
(\Cref{app:model_configs}). Per-task pass-rate changes between
TB2.0 and TB2.1 are reported in \Cref{app:per_task_changes}.

The workflow has also been incorporated into Terminal-Bench's
ongoing task-review process, documented at
\url{https://github.com/harbor-framework/terminal-bench/blob/main/docs/TASK_REVIEW_AUTOMATION.md}.
This implementation continues to evolve through contributions
from maintainers and the community and may differ from the
version evaluated in this paper.

To support reproducibility, the project website at
\url{https://ekbuchanan.com/terminal-bench-2-1-paper/}
provides a snapshot of the workflow, prompts, and results
reported in this paper.

\section*{Acknowledgements}

We thank the Terminal Bench Team and members of the community for reporting issues in the public GitHub repositories. We gratefully acknowledge the support of NIH under Nos.\ RF1MH133778, U54EB020405 (Mobilize), and R01GM124443 (SimTK); NSF under Nos. CCF2247015 (Hardware-Aware), CCF1763315 (Beyond Sparsity), CCF1563078 (Volume to Velocity), 2223827, 1937301 (RTML), 1900638 (Open Federated Virtual Assistants), 2028602 (PPoSS), and 24-554 (AIMing); DARPA under No. HR00112520038 (Fallingwater); US DEVCOM ARL under Nos. W911NF-23-2-0184 (Long-context) and W911NF-21-2-0251 (Interactive Human-AI Teaming); NSTXL under No. N00164-23-9-G057-01 (Energy-Efficient AI Hardware); ONR under Nos. N000142312633 (Deep Signal Processing), N000142212426 (Misinformation Analysis), N000142012480 (Non-Euclidean Geometry), and N000142012275 (Data Programming); Stanford HAI under Nos. 247183, 215955, and 345077 (Evo); Snorkel AI (Open Benchmarks Grant); Together AI; Google DeepMind; Google Research; Google Cloud; NXP, Xilinx, LETI-CEA, Intel, IBM (Code Generation), Microsoft, NEC, Toshiba, TSMC, ARM, Hitachi, BASF, Accenture, Ericsson, Qualcomm, Analog Devices, Salesforce, Total, Lightspeed, the Laude Institute, Prime Intellect, Anthropic, the HAI-GCP Cloud Credits for Research program, the Stanford Data Science Initiative (SDSI), the Stanford Marlowe Computing Platform, the McKnight Foundation, and members of the Stanford DAWN project: Meta, Google, and VMWare; and members of the Stanford SEAMS project: IBM and Felicis. This work was also partially supported by IBM and Felicis as an affiliate member of Stanford Institute for Human-centered Artificial Intelligence (HAI)'s industry program. We also gratefully acknowledge computing resources provided by Stanford's Marlowe Computing Instrument. The U.S. Government is authorized to reproduce and distribute reprints for Governmental purposes notwithstanding any copyright notation thereon. Any opinions, findings, and conclusions or recommendations expressed in this material are those of the authors and do not necessarily reflect the views, policies, or endorsements, either expressed or implied, of NIH, ONR, DARPA, or the U.S. Government.

\bibliography{references}

@misc{tb-repo,
  author       = {{Terminal-Bench Team}},
  title        = {Terminal-Bench: {GitHub} issue and pull-request tracker},
  year         = {2026},
  url          = {https://github.com/harbor-framework/terminal-bench},
  note         = {Accessed May~2026; community-reported defects against the public TB and TB2.0 task sets.}
}

@misc{tb2-repo,
  author       = {{Terminal-Bench Team}},
  title        = {Terminal-Bench 2.0: {GitHub} issue and pull-request tracker},
  year         = {2026},
  url          = {https://github.com/harbor-framework/terminal-bench-2},
  note         = {Accessed May~2026; community-reported defects against the public TB2.0 task set.}
}

@inproceedings{jain2024livecodebench,
  title     = {{LiveCodeBench}: Holistic and Contamination Free Evaluation of Large Language Models for Code},
  author = {Jain, Naman and Han, King and Gu, Alex and Li, Wen-Ding and Yan, Fanjia and Zhang, Tianjun and others},
  booktitle = {The Thirteenth International Conference on Learning Representations},
  year      = {2025},
  url       = {https://arxiv.org/abs/2403.07974}
}

@inproceedings{livebench,
  title={LiveBench: A Challenging, Contamination-Free {LLM} Benchmark},
  author={Colin White and Samuel Dooley and Manley Roberts and Arka Pal and Benjamin Feuer and Siddhartha Jain and Ravid Shwartz-Ziv and Neel Jain and Khalid Saifullah and Sreemanti Dey and Shubh-Agrawal and Sandeep Singh Sandha and Siddartha Venkat Naidu and Chinmay Hegde and Yann LeCun and Tom Goldstein and Willie Neiswanger and Micah Goldblum},
  booktitle={The Thirteenth International Conference on Learning Representations},
  year={2025},
}

@inproceedings{terminalbench,
  title     = {{Terminal-Bench}: Benchmarking Agents on Hard, Realistic Tasks in Command Line Interfaces},
  author    = {Merrill, Mike A. and Shaw, Alexander G. and Carlini, Nicholas and Li, Boxuan and Raj, Harsh and Bercovich, Ivan and others},
  booktitle = {International Conference on Learning Representations (ICLR)},
  year      = {2026},
  url       = {https://proceedings.iclr.cc/paper_files/paper/2026/hash/444a3737adaee10d86ad2ef5f74468e6-Abstract-Conference.html}
}

@article{chen2021codex,
  title   = {Evaluating large language models trained on code},
  author  = {Chen, Mark and Tworek, Jerry and Jun, Heewoo and Yuan, Qiming and others},
  journal = {arXiv preprint},
  year    = {2021},
  url     = {https://arxiv.org/abs/2107.03374}
}

@article{recht2019imagenet,
  title   = {Do {ImageNet} classifiers generalize to {ImageNet}?},
  author  = {Recht, Benjamin and Roelofs, Rebecca and Schmidt, Ludwig and Shankar, Vaishaal},
  journal = {Proceedings of ICML},
  year    = {2019},
  url     = {https://arxiv.org/abs/1902.10811}
}

@inproceedings{swebench,
  title     = {{SWE-bench}: Can Language Models Resolve Real-World {GitHub} Issues?},
  author    = {Jimenez, Carlos E. and Yang, John and Wettig, Alexander and Yao, Shunyu and Pei, Kexin and Press, Ofir and Narasimhan, Karthik},
  booktitle = {International Conference on Learning Representations (ICLR)},
  year      = {2024},
  url       = {https://arxiv.org/abs/2310.06770}
}

@misc{swebench-verified,
  title        = {Introducing {SWE-bench Verified}},
  author       = {{OpenAI}},
  year         = {2024},
  url          = {https://openai.com/index/introducing-swe-bench-verified/},
  howpublished = {OpenAI blog post},
  note         = {Human-validated 500-task subset of \citet{swebench}}
}

@inproceedings{webarena,
  title     = {{WebArena}: A Realistic Web Environment for Building Autonomous Agents},
  author = {Zhou, Shuyan and Xu, Frank F. and Zhu, Hao and Zhou, Xuhui and Lo, Robert and Sridhar, Abishek and others},
  booktitle = {International Conference on Learning Representations (ICLR)},
  year      = {2024},
  url       = {https://arxiv.org/abs/2307.13854}
}

@misc{webarena-verified,
  title        = {{WebArena Verified}: Reliable Evaluation for Web Agents},
  author       = {{ServiceNow Research}},
  year         = {2026},
  url          = {https://github.com/ServiceNow/webarena-verified},
  howpublished = {Software release},
  note         = {Audited 812-task subset of \citet{webarena}}
}

@misc{anthropic-infra-noise,
  title        = {Quantifying Infrastructure Noise in Agentic Coding Evals},
  author       = {Anthropic},
  year         = {2026},
  url          = {https://www.anthropic.com/engineering/infrastructure-noise},
  howpublished = {Anthropic Engineering Blog}
}

@inproceedings{react,
  title     = {{ReAct}: Synergizing Reasoning and Acting in Language Models},
  author    = {Yao, Shunyu and Zhao, Jeffrey and Yu, Dian and Du, Nan and Shafran, Izhak and Narasimhan, Karthik and Cao, Yuan},
  booktitle = {International Conference on Learning Representations (ICLR)},
  year      = {2023},
  url       = {https://arxiv.org/abs/2210.03629}
}

@misc{meng2025docent,
  title = {Introducing Docent},
  author = {Meng, Kevin and Huang, Vincent and Steinhardt, Jacob and Schwettmann, Sarah},
  year = {2025},
  url = {https://transluce.org/introducing-docent},
  note = {Transluce AI Technical Report}
}

@misc{langsmith,
  title        = {LangSmith: AI Agent and LLM Observability Platform},
  author       = {{LangChain}},
  year         = {2026},
  howpublished = {\url{https://www.langchain.com/langsmith/observability}},
  note         = {Accessed: 2026-06-30}
}

@misc{openai2025ghosts,
  title = {Ghosts in the Codex Machine},
  author = {{OpenAI}},
  year = {2025},
  howpublished = {Technical report},
  note = {\href{https://docs.google.com/document/d/1fDJc1e0itJdh0MXMFJtkRiBcxGEFtye6Xc6Ui7eMX4o}{OpenAI technical report}}
}

@inproceedings{yang2024sweagent,
  title = {{SWE}-agent: Agent-Computer Interfaces Enable Automated Software Engineering},
  author = {John Yang and Carlos E Jimenez and Alexander Wettig and Kilian Lieret and Shunyu Yao and Karthik R Narasimhan and Ofir Press},
  booktitle = {The Thirty-eighth Annual Conference on Neural Information Processing Systems},
  year = {2024},
  url = {https://arxiv.org/abs/2405.15793}
}

@misc{laude2025terminalbench,
  title = {Terminus: The Terminal-Bench Baseline Agent},
  author = {Merrill, Mike A. and Shaw, Alexander G. and others},
  year = {2025},
  note = {Baseline agent scaffold for Terminal-Bench},
  url = {https://github.com/terminal-bench/terminus}
}

@inproceedings{
wang2025openhands,
title={OpenHands: An Open Platform for {AI} Software Developers as Generalist Agents},
author={Xingyao Wang and Boxuan Li and Yufan Song and Frank F. Xu and Xiangru Tang and Mingchen Zhuge and Jiayi Pan and Yueqi Song and Bowen Li and Jaskirat Singh and Hoang H. Tran and Fuqiang Li and Ren Ma and Mingzhang Zheng and Bill Qian and Yanjun Shao and Niklas Muennighoff and Yizhe Zhang and Binyuan Hui and Junyang Lin and Robert Brennan and Hao Peng and Heng Ji and Graham Neubig},
booktitle={The Thirteenth International Conference on Learning Representations},
year={2025},
url={https://openreview.net/forum?id=OJd3ayDDoF}
}

@misc{openai2025codex,
  title = {Codex {CLI}: {A} lightweight coding agent that runs in the terminal},
  author = {{OpenAI}},
  year = {2025},
  howpublished = {Software platform},
  url = {https://openai.com/codex},
  note = {CLI reference: \url{https://developers.openai.com/codex/cli/reference/}}
}

@misc{anthropic2025agent,
  title = {Claude Code: An agentic coding tool},
  author = {{Anthropic}},
  year = {2025},
  howpublished = {Software documentation},
  url = {https://docs.anthropic.com/en/docs/agents-and-tools/claude-code/overview},
  note = {CLI reference: \url{https://docs.anthropic.com/en/docs/agents-and-tools/claude-code/cli-reference}}
}

@misc{google2025gemini,
  title = {Gemini {CLI}: An open-source {AI} agent for the terminal},
  author = {{Google}},
  year = {2025},
  howpublished = {Open-source software (Apache 2.0)},
  url = {https://github.com/google-gemini/gemini-cli},
  note = {Documentation: \url{https://geminicli.com/docs/}; tool reference: \url{https://geminicli.com/docs/tools/}}
}

@techreport{openai2026gpt53codexsystemcard,
  author      = {{OpenAI}},
  title       = {{GPT-5.3-Codex} System Card},
  institution = {OpenAI},
  year        = {2026},
  url         = {https://deploymentsafety.openai.com/gpt-5-3-codex/gpt-5-3-codex.pdf}
}

@techreport{openai2026gpt54systemcard,
  author      = {{OpenAI}},
  title       = {{GPT-5.4 Thinking} System Card},
  institution = {OpenAI},
  year        = {2026},
  url         = {https://deploymentsafety.openai.com/gpt-5-4-thinking/gpt-5-4-thinking.pdf}
}

@techreport{openai2026gpt55systemcard,
  author      = {{OpenAI}},
  title       = {{GPT-5.5} System Card},
  institution = {OpenAI},
  year        = {2026},
  url         = {https://deploymentsafety.openai.com/gpt-5-5/gpt-5-5.pdf}
}

@techreport{google2025gemini3flash,
  author      = {{Google DeepMind}},
  title       = {{Gemini 3 Flash}: Model Card},
  institution = {Google DeepMind},
  year        = {2025},
  month       = dec,
  url         = {https://storage.googleapis.com/deepmind-media/Model-Cards/Gemini-3-Flash-Model-Card.pdf}
}

@techreport{google2026gemini31pro,
  author      = {{Google DeepMind}},
  title       = {{Gemini 3.1 Pro}: Model Card},
  institution = {Google DeepMind},
  year        = {2026},
  month       = feb,
  url         = {https://storage.googleapis.com/deepmind-media/Model-Cards/Gemini-3-1-Pro-Model-Card.pdf}
}

@inproceedings{yu2025utboost,
  title     = {{UTBoost}: Rigorous Evaluation of Coding Agents on {SWE-Bench}},
  author    = {Yu, Boxi and others},
  booktitle = {Proceedings of the Annual Meeting of the Association for Computational Linguistics (ACL)},
  year      = {2025},
  url       = {https://arxiv.org/abs/2506.09289}
}

@inproceedings{kapoor2025hal,
  title     = {Holistic Agent Leaderboard: The Missing Infrastructure for {AI} Agent Evaluation},
  author    = {Kapoor, Sayash and Stroebl, Benedikt and Kirgis, Peter and
               Nadgir, Nitya and Siegel, Zachary S. and Wei, Boyi and
               Xue, Tianci and Chen, Ziru and others},
  booktitle = {International Conference on Learning Representations (ICLR)},
  year      = {2026},
  url       = {https://proceedings.iclr.cc/paper_files/paper/2026/hash/a0928f924a344aaebbb7f6cd8d56e34c-Abstract-Conference.html}
}

@article{kimiteam2026k25,
  title   = {{Kimi K2.5}: Visual Agentic Intelligence},
  author  = {{Kimi Team}},
  journal = {arXiv preprint arXiv:2602.02276},
  year    = {2026},
  url     = {https://arxiv.org/abs/2602.02276}
}

@misc{google2026gemma4,
  title        = {{Gemma 4}: Open Multimodal Models},
  author       = {{Google DeepMind}},
  year         = {2026},
  url          = {https://deepmind.google/models/gemma/gemma-4/},
  howpublished = {Model card and announcement;
                  \url{https://ai.google.dev/gemma/docs/core/model_card_4}},
  note         = {Released April 2026. Sizes: E2B, E4B, 26B MoE,
                  31B Dense; 256K context. We use the 31B Dense
                  variant in the broader 20-pair audit pool.}
}

@inproceedings{zhu2025abc,
  title     = {Establishing Best Practices in Building Rigorous Agentic Benchmarks},
  author    = {Zhu, Yuxuan and Jin, Tengjun and Pruksachatkun, Yada and Zhang, Andy and Liu, Shu and Cui, Sasha and others},
  booktitle = {Advances in Neural Information Processing Systems},
  volume    = {38},
  year      = {2025},
}

@misc{anthropic2026opus46,
  author       = {{Anthropic}},
  title        = {Introducing {Claude Opus 4.6}},
  howpublished = {Anthropic announcement and system card},
  year         = {2026},
  month        = {February 5},
  note         = {Terminal-Bench 2.0 reported on Terminus-2 harness (Codex CLI for OpenAI models), 1$\times$ guaranteed / 3$\times$ ceiling resource allocation, 5--15 samples per task across staggered batches.},
  url          = {https://www.anthropic.com/news/claude-opus-4-6}
}

@misc{anthropic2026sonnet46,
  author       = {{Anthropic}},
  title        = {Introducing {Claude Sonnet 4.6}},
  howpublished = {Anthropic announcement and system card},
  year         = {2026},
  month        = {February 17},
  url          = {https://www.anthropic.com/news/claude-sonnet-4-6}
}

@misc{qwen2026qwen359b,
  author       = {{Qwen Team}},
  title        = {{Qwen3.5-9B} Model Card},
  howpublished = {Hugging Face model card},
  year         = {2026},
  note         = {Revision c202236235762e1c871ad0ccb60c8ee5ba337b9a},
  url          = {https://huggingface.co/Qwen/Qwen3.5-9B/blob/c20223623576/README.md}
}

@misc{qwen2026qwen3527b,
  author       = {{Qwen Team}},
  title        = {{Qwen3.5-27B} Model Card},
  howpublished = {Hugging Face model card},
  year         = {2026},
  note         = {Revision fc05daec18b0a78c049392ed2e771dde82bdf654},
  url          = {https://huggingface.co/Qwen/Qwen3.5-27B/blob/fc05daec18b0/README.md}
}

@misc{zai-tb2-verified,
  author       = {{Z.ai}},
  title        = {Terminal-Bench 2.0 Verified: Instruction \& Environment Fix Version},
  year         = {2026},
  howpublished = {Hugging Face dataset, \url{https://huggingface.co/datasets/zai-org/terminal-bench-2-verified}},
  note         = {Documents 89 environment fixes and 11 instruction fixes;
                  reports a $5.5$~pp shift on Claude Opus~4.5 from environment fixes alone.
                  Companion to GLM-5~\citep{zeng2026glm5}.}
}

@article{zeng2026glm5,
  title   = {{GLM-5}: from Vibe Coding to Agentic Engineering},
  author = {Zeng, Aohan and Lv, Xin and Hou, Zhenyu and Du, Zhengxiao and
             Zheng, Qinkai and Chen, Bin and Yin, Da and others},
  journal = {arXiv preprint},
  year    = {2026},
  url     = {https://arxiv.org/abs/2602.15763}
}

@misc{openai2026harness,
  title = {Harness engineering: leveraging {Codex} in an agent-first world},
  author = {{OpenAI}},
  year = {2026},
  month = feb,
  howpublished = {\url{https://openai.com/index/harness-engineering/}},
  note = {Accessed: 2026-05-05}
}

@misc{shi2026harboradaptersharborindexinfrastructure,
      title={Harbor Adapters and Harbor-Index: Infrastructure and a Curated Meta-Dataset for Large-Scale Agentic Evaluation}, 
      author={Lin Shi and Haowei Lin and Zixuan Zhu and Xiaoyue Zhou and Xiang Li et al.},
      year={2026},
      eprint={2609.04298},
      archivePrefix={arXiv},
      primaryClass={cs.AI},
      url={https://arxiv.org/abs/2609.04298}, 
}
\bibliographystyle{plainnat}

\appendix 
\onecolumn
\appendix
\raggedbottom
\crefalias{section}{appendix}
\crefalias{subsection}{appendix}

\etocdepthtag.toc{app}
\begingroup
  \etocsettagdepth{main}{none}
  \etocsettagdepth{app}{subsection}
  \etocsettocstyle{\section*{Appendix Contents}}{}
  \setcounter{tocdepth}{2}
  \hypersetup{linkcolor=black}
  \tableofcontents
\endgroup
\bigskip

\bigskip
\bigskip

\clearpage

\clearpage
\section{Audit Workflow Details}
\label{app:audit_pipeline}

\subsection{Continuous-validation pipeline}
\label{app:continuous_validation}
As benchmark users continue to open issues and pull requests, maintainers need a lightweight way to check whether proposed changes preserve task validity. We implement this audit workflow with GitHub Actions that can be invoked manually or triggered automatically on pull requests for a given task. 

The pipeline exposes two checks:
\begin{itemize}
\item \texttt{/audit} calls the auditor agent to review the task instructions and available trial logs.
\item \texttt{/oracle} builds the task environment and runs the reference solution, checking that a known-correct solution still passes.
\end{itemize}

\paragraph{Propose-only.}
Although agents could be given permission to propose fixes and merge pull requests automatically, we deliberately restrict the workflow to propose-only operation. Each run is capped at five repair iterations, and the workflow grants pull-requests but prevents agents from directly changing repository state. Instead, proposed fixes are emitted as maintainer-reviewable diffs in \texttt{audit/changes/<task>.diff}.

The audit rubrics are integrated in \href{https://github.com/harbor-framework/terminal-bench}{https://github.com/harbor-framework/terminal-bench}.

The original discussion can be found in \href{https://github.com/harbor-framework/terminal-bench-2/pull/53}{https://github.com/harbor-framework/terminal-bench-2/pull/53}

\subsection{Audit Workflow Evaluation}
\label{app:judge_prompts}

We measure agreement at two levels. Managed coding assistants, Claude
Code (Opus~4.6) and Codex (GPT-5.4), agree on the primary defect
category, including \emph{no defect}, at $\kappa=0.80$ over all 89 tasks. Lower-cost API auditors are
evaluated on the binary triage decision (flagged vs.\ clean) over all
89 TB2.0 tasks (\Cref{tab:canary_agreement}), with pairwise agreement
of $\kappa=0.55$--$0.77$. Per category (one-vs-rest), their agreement
is moderate for instruction--verification mismatch ($\kappa=0.60$) and
environment issues ($\kappa=0.59$) and low for task drift
($\kappa=0.21$--$0.41$). Because every drift defect breaks the
reference solution, the deterministic \texttt{/oracle} check catches
drift regardless of auditor labels.

\begin{table}[H]
\centering
\caption[Inter-judge agreement.]{
  \textbf{Inter-judge agreement.} Cohen's $\kappa$ between pairs of
  judges on the binary flag (flagged vs. audit-clean) over all 89 TB2.0
  tasks.
}
\label{tab:canary_agreement}
\footnotesize
\setlength{\tabcolsep}{4pt}
\begin{tabular}{llrrr}
\toprule
Agent A & Agent B & Agree & Disagree & $\kappa$ \\
\midrule
gpt-4o-mini & gpt-4o & 69 & 20 & 0.550 \\
gpt-4o-mini & gemini-2.5-flash & 79 & 10 & 0.769 \\
gpt-4o & gemini-2.5-flash & 69 & 20 & 0.549 \\
\bottomrule
\end{tabular}
\end{table}

\paragraph{Per-task map to error categories}
\cref{tab:failure_categories} shows the failure modes assigned to different tasks found to need repair.
\Cref{tab:revised_by_stage} shows the final $28$ revised tasks, and which component in the task was updated for TB~2.1.

\begin{table*}[t]
  \centering
  \footnotesize
  \caption{%
    \textbf{Failure-modes among the 28 tasks revised
    in TB2.1.}
  }
  \setlength{\tabcolsep}{6pt}
  \begin{tabular}{p{0.30\textwidth} p{0.30\textwidth} p{0.30\textwidth}}
    \toprule
    \textbf{Instruction--verification mismatch (16)} & \textbf{Environment issues (8)} & \textbf{Task drift (12)} \\
    \midrule
    adaptive-rejection-sampler, build-pmars, caffe-cifar-10,
    configure-git-webserver, filter-js-from-html, fix-git,
    hf-model-inference, install-windows-3.11, mteb-leaderboard,
    mteb-retrieve, polyglot-c-py, polyglot-rust-c, query-optimize,
    sam-cell-seg, torch-tensor-parallelism, train-fasttext
    &
    caffe-cifar-10, compile-compcert, crack-7z-hash,
    filter-js-from-html, gpt2-codegolf, query-optimize,
    torch-pipeline-parallelism, torch-tensor-parallelism
    &
    adaptive-rejection-sampler, build-pmars, build-pov-ray,
    extract-moves-from-video, financial-document-processor,
    make-doom-for-mips, mcmc-sampling-stan, mteb-leaderboard,
    mteb-retrieve, overfull-hbox, protein-assembly, rstan-to-pystan
    \\
    \bottomrule
  \end{tabular}
  \label{tab:failure_categories}
\end{table*}

\begin{table*}[ht]
  \centering
  \scriptsize
  \caption{%
    \textbf{Breakdown of the 28 TB~2.0 tasks revised in TB~2.1, by the stage of the task definition that required repair.}
    }
      \setlength{\tabcolsep}{4pt}
  \begin{tabular}{lccccccc}
    \toprule
    \textbf{Task} & \textbf{Instr.} & \textbf{Env.} & \textbf{Oracle} & \textbf{Tests} & \textbf{Time} & \textbf{Mem.} & \textbf{CPUs} \\
    \midrule
    adaptive-rejection-sampler   & \checkmark &            & \checkmark &            &            &            &            \\
    build-pmars                  & \checkmark &            &            & \checkmark &            &            &            \\
    build-pov-ray                &            &            & \checkmark &            &            &            &            \\
    caffe-cifar-10               & \checkmark & \checkmark &            & \checkmark & \checkmark & \checkmark & \checkmark \\
    crack-7z-hash                &            &            &            &            & \checkmark & \checkmark &            \\
    compile-compcert             &            & \checkmark &            &            &            &            & \checkmark \\
    configure-git-webserver      &            &            & \checkmark & \checkmark &            &            &            \\
    extract-moves-from-video     & \checkmark & \checkmark &            &            &            &            &            \\
    filter-js-from-html          & \checkmark &            &            &            & \checkmark & \checkmark &            \\
    financial-document-processor &            &            & \checkmark &            &            &            &            \\
    fix-git                      &            & \checkmark &            & \checkmark &            &            &            \\
    gpt2-codegolf                &            &            &            &            &            & \checkmark &            \\
    hf-model-inference           &            &            &            & \checkmark &            &            &            \\
    install-windows-3.11         & \checkmark &            &            & \checkmark &            &            &            \\
    make-doom-for-mips           &            &            & \checkmark &            &            &            &            \\
    mcmc-sampling-stan           &            &            &            & \checkmark &            &            &            \\
    mteb-leaderboard             & \checkmark & \checkmark &            &            &            & \checkmark &            \\
    mteb-retrieve                & \checkmark & \checkmark &            &            &            &            &            \\
    overfull-hbox                &            &            & \checkmark &            &            &            &            \\
    polyglot-c-py                &            &            &            & \checkmark &            &            &            \\
    polyglot-rust-c              &            &            &            & \checkmark &            &            &            \\
    protein-assembly             &            & \checkmark & \checkmark &            &            &            &            \\
    query-optimize               & \checkmark &            &            &            & \checkmark &            &            \\
    rstan-to-pystan              &            &            & \checkmark &            &            &            &            \\
    sam-cell-seg                 & \checkmark &            &            & \checkmark &            &            &            \\
    torch-pipeline-parallelism   &            &            &            &            &            & \checkmark &            \\
    torch-tensor-parallelism     & \checkmark &            &            &            &            & \checkmark &            \\
    train-fasttext               &            &            &            & \checkmark &            &            &            \\
    \bottomrule
  \end{tabular}
  \label{tab:revised_by_stage}
\end{table*}

\subsection{Comparison with Terminal-Bench 2.0 Verified}
\label{app:tb_zai}

Z.ai released Terminal-Bench 2.0 Verified~\citep{zai-tb2-verified}, an earlier public audit of TB2.0.  Terminal-Bench 2.0 Verified makes two kinds of changes. First, environment fixes applied to all 89 tasks make the task images run under the Claude Code runtime: for example, installing \texttt{procps}, whose absence crashed the agent, and capping the cores visible to \texttt{caffe-cifar-10} so that parallel builds no longer run out of memory. Z.ai reports that these fixes alone raise Claude Opus~4.5 from 52.4\% to 57.9\%, an early demonstration that environment defects, not only task content, move leaderboard scores. Second, instruction fixes correct 11 tasks whose instructions and tests disagree.

Terminal-Bench~2.1 repairs these 11 tasks. It adopts Verified's repair on four tasks. 

We note two core differences. First, when an instruction and its test disagree, Verified edits the instruction to state what the test checks, whereas Terminal-Bench~2.1 more often edits the test to accept any correct solution; stating the grader's expectation tells the agent part of the solution, while relaxing the grader leaves the task as hard as its contributor intended. Second, Verified adapts the task images to a specific harness, including a note on five tasks telling the agent not to use the Read tool on images or PDFs:

\begin{quote}
\small
\ttfamily
\raggedright
IMPORTANT: You do not have multimodal capabilities. Do not use the Read tool to directly read images or PDFs. Please use other approaches to process the content.%
\end{quote}

\noindent TB2.1 adds no harness-specific changes, so every agent is evaluated on the same task.

\section{Terminal-Bench Benchmark Evaluation}
\label{app:eval_setup}

\subsection{Models evaluated}
\label{app:models}

The evidence data included leaderboard trials from the 13 models listed in \Cref{tab:models}. %

\begin{table*}[ht]
  \centering
  \footnotesize
  \caption{%
    \textbf{Models evaluated.} One row per model entry
    used in the agent-rank comparison or compute-scaling analyses.
    \textsc{ctx}~values in italics are pinned smaller by the harness
    config; plain values are the vendor maximum.
  }
  \setlength{\tabcolsep}{6pt}
  \begin{tabular}{l l l r}
    \toprule
    \textbf{Model} & \textbf{Provider} & \textbf{Route} & \textbf{Ctx} \\
    \midrule
    Claude Opus 4.6 \citep{anthropic2026opus46}                      & Anthropic   & Anthropic API   & 200K \\
    Claude Sonnet 4.6 \citep{anthropic2026sonnet46}                  & Anthropic   & Anthropic API   & 200K \\
    \midrule
    GPT 5.3 Codex \citep{openai2026gpt53codexsystemcard}             & OpenAI      & OpenAI API      & 400K \\
    GPT 5.4 \citep{openai2026gpt54systemcard}                        & OpenAI      & OpenAI API      & 400K \\
    GPT 5.4 mini \citep{openai2026gpt54systemcard}                   & OpenAI      & OpenAI API      & 400K \\
    GPT 5.5 \citep{openai2026gpt55systemcard}                        & OpenAI      & OpenAI API      & 400K \\
    \midrule
    Gemini Flash 3 \citep{google2025gemini3flash}                    & Google      & Google API      & 1M \\
    Gemini Pro 3.1 \citep{google2026gemini31pro}                     & Google      & Google API      & 1M \\
    Gemma 4 31B \citep{google2026gemma4}                             & Google      & Together.ai     & 256K \emph{(131K)} \\
    \midrule
    GLM 5.1 \citep{zeng2026glm5}                                     & Z.AI        & Anthropic-shape & 200K \\
    \midrule
    Kimi K2.6 \citep{kimiteam2026k25}                                & Moonshot AI & Together.ai     & 256K \emph{(131K)} \\
    Qwen 3.5 9B \citep{qwen2026qwen359b}                             & Alibaba     & Together.ai     & 254K \emph{(131K)} \\
    Qwen 3.5 27B \citep{qwen2026qwen3527b}                           & Alibaba     & Together.ai     & 254K \\
    \bottomrule
  \end{tabular}
  \label{tab:models}
\end{table*}

\subsection{Harness Evaluation}
\label{app:harnesses}
Models were evaluated with a common baseline harness, \texttt{terminus-2} \citep{terminalbench}, and, where available, the model provider's preferred agent harness. The latter includes Anthropic's Claude Code CLI \citep{anthropic2025agent}, OpenAI's Codex CLI
\citep{openai2025codex,openai2026harness}, and Google's Gemini CLI
\citep{google2025gemini}.

\begin{table*}[t]
  \centering
  \scriptsize
  \caption{%
    \textbf{Harnesses evaluated.}
  }
  \setlength{\tabcolsep}{4pt}
  \renewcommand{\arraystretch}{1.15}
  \begin{tabularx}{\textwidth}{@{} l l X @{}}
    \toprule
    \textbf{Harness} & \textbf{Scaffold} &
    \textbf{Source / defaults} \\
    \midrule
    \texttt{terminus-2}  & single-agent ReAct~\citep{react}
      & Terminal-Bench 2.0 baseline~\citep{laude2025terminalbench,tb-repo};
        including JSON-action parser, shell + file I/O tools.
        \textbf{Defaults:} \texttt{enable\_summarize=True},
        \texttt{proactive\_summarization\_threshold=8000},
        \texttt{temperature=1.0} for thinking-capable models. \\
    \midrule
    \texttt{claude-code} & provider-managed agentic CLI
      & Anthropic Claude Code, headless~\citep{anthropic2025agent};
        self-managed tool palette + context compaction.
        \textbf{Defaults in our runs:} \texttt{reasoning\_effort=high},
        adaptive thinking on. \\
    \midrule
    \texttt{codex}       & provider-managed agentic CLI
      & OpenAI Codex CLI, non-interactive~\citep{openai2025codex,openai2026harness};
        native Responses-API tool calls, managed context.
        \texttt{use\_responses\_api=True}. \\
    \midrule
    \texttt{gemini-cli}  & provider-managed agentic CLI
      & Google Gemini CLI, Apache~2.0~\citep{google2025gemini}; shell +
        file I/O tools, managed context. \\
    \bottomrule
  \end{tabularx}
  \label{tab:harnesses}
\end{table*}

\subsection{Agent configuration overrides}
\label{app:model_configs}

\Cref{tab:config_deltas} lists the agent configurations. Rows marked ``default'' use the
harness-level defaults documented in \Cref{tab:harnesses}.

We note that \texttt{terminus-2} triggers proactive summarization when the free context drops below a threshold (default 8K tokens). We leave the default for frontier-API models and GLM~5.1, where a single turn rarely exceeds 8K tokens. For the open-weight long-context models
(Gemma~4~31B and Kimi~K2.6, both with 256K context windows), we raise the threshold to 131K tokens after observing trials where summarization fired too late to leave room for one more turn. 

The Gemma~4~31B and Kimi~K2.6 rows in \Cref{tab:config_deltas} are
collapsed for readability; the full sampling settings are:
\begin{itemize}\setlength{\itemsep}{1pt}
  \item \textbf{Gemma~4~31B:} temperature $1.0$, interleaved
    thinking on, max input tokens $253{,}952$, max output tokens
    $8{,}192$, proactive-summarization threshold $131{,}072$,
    \texttt{top\_p}$=0.95$, \texttt{top\_k}$=20$,
    \texttt{presence\_penalty}$=1.5$.
  \item \textbf{Kimi~K2.6:} temperature $0.6$, interleaved
    thinking on, max input tokens $253{,}952$, max output tokens
    $8{,}192$, proactive-summarization threshold $131{,}072$,
    \texttt{top\_p}$=0.95$, \texttt{top\_k}$=20$.
\end{itemize}

\begin{table*}[t]
  \centering
  \footnotesize
  \caption{%
    \textbf{Agent--model configurations.}
    ``Default'' means the harness-level defaults of
    \Cref{tab:harnesses} apply unchanged. 
    Additional overrides are listed below.
  }
  \setlength{\tabcolsep}{5pt}
  \renewcommand{\arraystretch}{1.1}
  \begin{tabularx}{\textwidth}{@{} l l X @{}}
    \toprule
    \textbf{Agent} & \textbf{Model} & \textbf{Overrides vs.\ harness default} \\
    \midrule
    claude-code & Claude Opus 4.6     & \texttt{reasoning\_effort=high} \\
    claude-code & Claude Sonnet 4.6   & \texttt{reasoning\_effort=high} \\
    claude-code & GLM 5.1             & default \\
    \midrule
    codex       & GPT 5.3 Codex       & \texttt{reasoning\_effort=xhigh} \\
    codex       & GPT 5.4             & \texttt{reasoning\_effort=xhigh} \\
    codex       & GPT 5.4 mini        & \texttt{reasoning\_effort=xhigh} \\
    codex       & GPT 5.5             & \texttt{reasoning\_effort=xhigh}, \texttt{temperature=1.0} \\
    \midrule
    gemini-cli  & Gemini Flash 3      & default \\
    gemini-cli  & Gemini Pro 3.1      & default \\
    \midrule
    terminus-2  & Claude Opus 4.6     & \texttt{reasoning\_effort=high}, \texttt{T=1} \\
    terminus-2  & Claude Sonnet 4.6   & \texttt{reasoning\_effort=high}, \texttt{T=1} \\
    terminus-2  & GLM 5.1             & \texttt{T=1.0}, \texttt{thinking.type=enabled} \\
    terminus-2  & GPT 5.3 Codex       & \texttt{reasoning\_effort=xhigh}, \texttt{use\_responses\_api=True} \\
    terminus-2  & GPT 5.4             & \texttt{reasoning\_effort=xhigh}, \texttt{use\_responses\_api=True} \\
    terminus-2  & GPT 5.4 mini        & \texttt{reasoning\_effort=xhigh}, \texttt{use\_responses\_api=True} \\
    terminus-2  & GPT 5.5             & \texttt{reasoning\_effort=xhigh}, \texttt{T=1}, \texttt{use\_responses\_api=True} \\
    terminus-2  & Gemini Flash 3      & \texttt{reasoning\_effort=high} \\
    terminus-2  & Gemini Pro 3.1      & \texttt{reasoning\_effort=high} \\
    terminus-2  & Gemma 4 31B         & \texttt{T=1.0}, \texttt{interleaved\_thinking=True}; open-weight pin block (see text) \\
    terminus-2  & Kimi K2.6           & \texttt{T=0.6}, \texttt{interleaved\_thinking=True}; open-weight pin block (see text) \\
    \bottomrule
  \end{tabularx}
  \label{tab:config_deltas}
\end{table*}

\clearpage
\section{Supplementary results}
\label{app:supp_results}
\label{app:continuous_validation_results}  %

\subsection{Per-task pass-rate changes (TB2.0 \texorpdfstring{$\to$}{to} TB2.1)}
\label{app:per_task_changes}

\Cref{tab:changed_tasks} reports the per-task pass-rate change from TB2.0 to TB2.1, averaged across all agents. After the changes no task remains unsolved. The largest regression is in task \texttt{overfull-hbox}:
 in TB2.0 a broken environment build left the verifier without ground truth, crediting outputs that the repaired verifier rejects.

\begin{table}[H]
  \centering
  \small
  \caption{%
    \textbf{Changed-task pass rates.}
    Rates aggregated across all agent model pairs evaluated. %
  }
  \setlength{\tabcolsep}{6pt}
  \begin{tabular}{lcc}
    \toprule
    \textbf{Task} & \textbf{TB2.0} & \textbf{TB2.1} \\
    \midrule
    adaptive-rejection-sampler   & 31.7\%             & 58.3\%             \\
    build-pmars                  & 75.0\%             & 81.7\%             \\
    build-pov-ray                & 75.0\%             & 68.3\%             \\
    caffe-cifar-10               & 5.0\%              & 69.2\%             \\
    compile-compcert             & 45.0\%             & 48.3\%             \\
    configure-git-webserver      & 46.7\%             & 50.0\%             \\
    crack-7z-hash                & 75.0\%             & 86.7\%             \\
    extract-moves-from-video     & 8.3\%              & 13.3\%             \\
    filter-js-from-html          & 0.0\%              & 6.7\%              \\
    financial-document-processor & 61.7\%             & 56.7\%             \\
    fix-git                      & 98.3\%             & 95.0\%             \\
    gpt2-codegolf                & 15.0\%             & 18.3\%             \\
    hf-model-inference           & 88.3\%             & 86.7\%             \\
    install-windows-3.11         & 0.0\%              & 23.3\%             \\
    make-doom-for-mips           & 6.7\%              & 1.7\%              \\
    mcmc-sampling-stan           & 75.0\%             & 71.7\%             \\
    mteb-leaderboard             & 28.3\%             & 36.7\%             \\
    mteb-retrieve                & 15.0\%             & 41.7\%             \\
    overfull-hbox                & 53.3\%             & 36.7\%             \\
    polyglot-c-py                & 5.0\%              & 80.0\%             \\
    polyglot-rust-c              & 5.0\%              & 66.7\%             \\
    protein-assembly             & 20.0\%             & 15.0\%             \\
    query-optimize               & 50.0\%             & 43.3\%             \\
    rstan-to-pystan              & 58.3\%             & 58.3\%             \\
    sam-cell-seg                 & 0.0\%              & 53.3\%             \\
    torch-pipeline-parallelism   & 8.3\%              & 10.0\%             \\
    torch-tensor-parallelism     & 15.0\%             & 66.7\%             \\
    train-fasttext               & 5.0\%              & 3.3\%              \\
    \bottomrule
  \end{tabular}
  \label{tab:changed_tasks}
\end{table}

\subsection{Agent Failure Decomposition}
\label{app:failure_composition}
Agentic evaluation infrastructure is still a work in progress: runner
failures, provider interruptions, authentication errors, sandbox crashes, and
setup timeouts can all occur before the agent has produced a task-level
outcome. To avoid mixing these infrastructure failures with model behavior,
this analysis uses a conservative whitelist-filtering approach. The table
builder retains only trials whose terminal state can be interpreted as an
agent outcome: a pass, a completed but wrong answer, or an
\texttt{AgentTimeoutError}. Timed-out trials are still graded on the state the agent left, and 11.8\% of graded timeouts pass. Other upstream infrastructure errors are filtered before the decomposition is computed. However, we note that these filtered infrastructure trials account for roughly $27\%$ of upstream trials.

\begin{figure*}[t]
  \centering
  \includegraphics[width=0.72\textwidth]{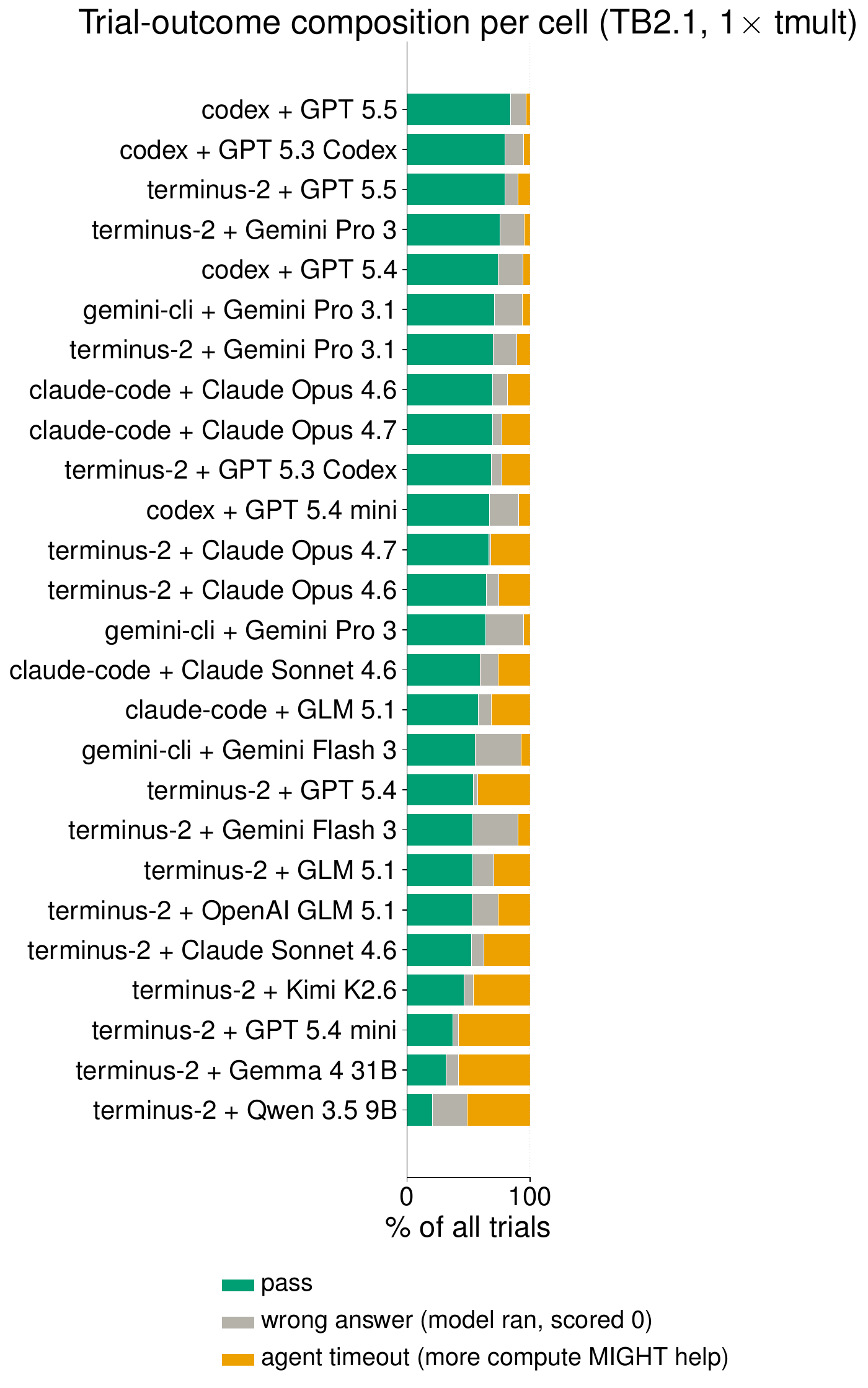}
  \caption{%
    \textbf{Trial outcomes by agent.} Each horizontal bar shows, for one agent, the
    fraction of trials that passed (green), returned a clean wrong
    answer (gray), or hit \texttt{AgentTimeoutError} (amber). 
  }
  \label{fig:failure_composition}
\end{figure*}

\Cref{fig:failure_composition} decomposes each whitelisted trial into one of
three mutually exclusive outcomes: a pass ($\texttt{reward}=1$), a completed
but incorrect answer ($\texttt{reward}=0$), or an agent timeout (\texttt{Agent\allowbreak Timeout\allowbreak Error}). At fixed budget, the weakest agents are dominated by timeouts rather than by scored wrong answers. This timeout-dominant pattern appears for \texttt{terminus-2} paired with Qwen~3.5~9B, Gemma~4~31B, Kimi~K2.6, and GPT~5.4~mini. For these cells, many runs end before the agent finishes. These results motivate the budget sweeps in
\Cref{sec:compute_scaling}.

\end{document}